\documentclass[conference]{IEEEtran}
\IEEEoverridecommandlockouts

\usepackage{cite}
\usepackage{amsmath,amssymb,amsfonts}
\usepackage[T1]{fontenc}
\usepackage{algorithmic}
\usepackage{graphicx}
\usepackage{textcomp}
\usepackage{xcolor}
\usepackage{subcaption}
\usepackage{makecell}
\usepackage{array}
\usepackage{booktabs}
\usepackage{eso-pic} 
\usepackage[letterpaper]{geometry}
\usepackage{float}
\def\BibTeX{{\rm B\kern-.05em{\sc i\kern-.025em b}\kern-.08em
    T\kern-.1667em\lower.7ex\hbox{E}\kern-.125emX}}

\newgeometry{top=25.4mm, bottom=19.1mm, left=19.1mm, right=19.1mm}
\begin{document}

\title{Multi-Agent Scenario Generation in Roundabouts with a Transformer-enhanced Conditional Variational Autoencoder}

\author{
    Li Li\textsuperscript{1}, 
    Tobias Brinkmann\textsuperscript{1}, 
    Till Temmen\textsuperscript{1}, 
    Markus Eisenbarth\textsuperscript{1},\\
    and Jakob Andert\textsuperscript{1}
}

\maketitle

\footnotetext[1]{Teaching and Research Area for Mechatronics in Mobile Propulsion, RWTH Aachen University, Germany, \{li\_li, brinkmann, temmen, eisenbarth, andert\}@mmp.rwth-aachen.de
}

\AddToShipoutPicture*{%
    \AtTextLowerLeft{%
        \raisebox{-3\baselineskip}[0pt][0pt]{%
            \makebox[\textwidth][c]{\footnotesize
            © 2025 IEEE. This is the author’s version of the paper accepted for publication in IEEE ITSC 2025.}%
        }%
    }%
}

\begin{abstract}
With the increasing integration of intelligent driving functions into serial-produced vehicles, ensuring their functionality and robustness poses greater challenges. Compared to traditional road testing, scenario-based virtual testing offers significant advantages in terms of time and cost efficiency, reproducibility, and exploration of edge cases. We propose a Transformer-enhanced Conditional Variational Autoencoder (CVAE-T) model for generating multi-agent traffic scenarios in roundabouts, which are characterized by high vehicle dynamics and complex layouts, yet remain relatively underexplored in current research. The results show that the proposed model can accurately reconstruct original scenarios and generate realistic, diverse synthetic scenarios. Besides, two Key-Performance-Indicators (KPIs) are employed to evaluate the interactive behavior in the generated scenarios. Analysis of the latent space reveals partial disentanglement, with several latent dimensions exhibiting distinct and interpretable effects on scenario attributes such as vehicle entry timing, exit timing, and velocity profiles. The results demonstrate the model’s capability to generate scenarios for the validation of intelligent driving functions involving multi-agent interactions, as well as to augment data for their development and iterative improvement.
\end{abstract}

\section{Introduction}

Intelligent driving functions are shaping the automotive industry. Functions like Adaptive Cruise Control (ACC) or Lane Keeping Assistance (LKA) can take the longitudinal or lateral control or both, thereby reducing driver workload and improving comfort. In order to enhance the driving safety, active safety systems have demonstrated significant potential in mitigating accident risks \cite{aleksa2024impact}. Automated Emergency Braking (AEB) and Front Collision Warning (FCW) can reduce up to 29\% car crashes and 9\% injuries. Lane Departure Warning (LDW) has shown the capability to reduce crash rates by approximately 7\% \cite{benson2018adas}.  

To ensure the reliability and robustness of these functionalities, comprehensive validation is required. Road testing in real-world environments is a conventional approach \cite{koopman2016challenges}. However, it is time- and cost-intensive, particularly when attempting to accumulate the extensive mileage required for safety validation \cite{wachenfeld2016release}. Moreover, real-world testing poses inherent safety risks to drivers due to the potential malfunctioning of unproven systems \cite{liu2018ar}. An increasingly adopted method is scenario-based testing in virtual environments. In \cite{ulbrich2015scene} a scenario can be defined as temporal sequence of maneuvers and environmental elements, including traffic elements, natural elements, road elements, etc. Scenarios are systematically designed to emulate real-world driving conditions and can be customized with particular emphasis on the trajectories and interactive behaviors of the traffic participants. In addition, scenario-based testing enables the exploration of rare and safety-critical traffic situations that are difficult to observe in real-world data \cite{jian2022scenario}. 

The generation of scenarios consisting of single-agent trajectory in relatively simple road layouts, such as highways, has been extensively studied. However, such approaches are insufficient for validating functions that involve interactions between multiple traffic participants, such as AEB, FCW or Collision Avoidance (CA). To address this limitation, traffic simulation involving multiple agents has attracted growing research attention. Nevertheless, these simulations often lack sufficient granularity in modeling the detailed long-term trajectories and interactive behavior among agents \cite{di2024data}. Realistic scenario generation that represents the spatio-temporal relationships between multiple agents remains a significant challenge, particularly in complex intersections where high vehicle dynamics and diverse maneuver possibilities introduce substantial behavioral uncertainty.

Roundabouts, as a type of at-grade intersection, have been widely implemented worldwide, especially in European countries such as France and UK \cite{brilon2021state}. In roundabouts, vehicles already inside typically have the right of way. Compared to vehicle maneuvers on highways, driving behavior within roundabouts is more complex. Entering and exiting a roundabout often involves high acceleration or deceleration, along with substantial steering input over a sustained duration due to the curved geometry.

In this paper, we propose a deep generative model for roundabout scenario generation using a Transformer-enhanced Conditional Variational Autoencoder (CVAE-T). In Section III, we utilize the rounD dataset \cite{krajewski2020round} to construct scenarios by extracting trajectories of two interacting vehicles within a same temporal window. A Transformer-enhanced CVAE model with variable $\beta$ scheduling strategy is developed to capture the spatio-temporal patterns of the vehicles and their interaction. In Section IV, experimental results are presented by comparing original scenarios and reconstructed scenarios. Two Key-Performance-Indicators (KPIs) are employed to assess the interactive behavior between vehicles. Finally, the disentanglement and interpretability of the learned latent space are analyzed.

In summary, the main contributions of this paper are as follows:

\begin{itemize}

\item We propose a data processing pipeline to extract scenario data and condition data from the dataset for CVAE training. The method is extensible to road layouts beyond roundabouts.

\item We propose a Transformer-enhanced Conditional Variational Autoencoder to capture complex spatio-temporal patterns of interacting vehicles in roundabouts.

\item We evaluate the model by comparing the distribution of two KPIs in original and generated scenarios. Furthermore, a detailed analysis of the learned latent space demonstrates the model’s capability to generate realistic, diverse and condition-consistent synthetic scenarios.
\end{itemize}

\section{Related work}

\subsection{Roundabout Dataset}\label{AA}

Roundabouts are common traffic elements in urban driving environments, posing unique challenges for trajectory planning and vehicle control. In recent years, several open-source datasets have been released with a specific focus on roundabout scenarios. The INTERACTION dataset provides traffic data from five different roundabouts at a  sampling frequency of 10 Hz, comprising 10479 vehicle trajectories in total over more than 360 minutes \cite{zhan2019interaction}. Another dataset with three roundabouts in Germany were recorded using a drone, capturing over 12700 traffic participants at a sampling frequency of 25 Hz. The resulting dataset, known as the rounD dataset, also includes high-definition maps provided in the OpenDRIVE format \cite{krajewski2020round}. The AA dataset contains detailed records of both normal and safety-critical driving scenarios with the tracking frequency 2.5 Hz\cite{yan2023realism}. These datasets offer opportunities and convenience for analyzing scenarios in roundabouts.

\begin{figure}[htbp]
\centerline{\includegraphics{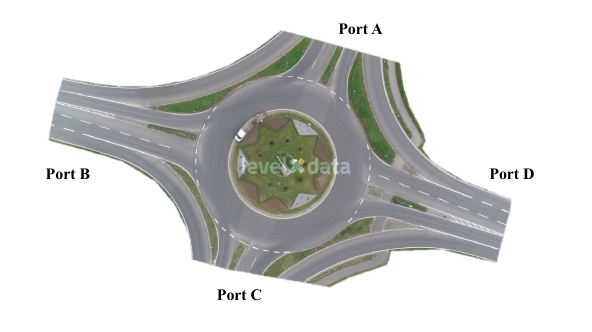}}
\caption{Neuweiler roundabout in rounD dataset \cite{krajewski2020round}}
\label{roundabout}
\end{figure}

\subsection{Deep Generative Models}

Scenario generation approaches generally fall into two categories: knowledge-based and data-driven methods \cite{nalic2020scenario} \cite{menzel2019functional}. Knowledge-based approaches offer high interpretability and realism, leveraging expert-defined rules to construct scenarios. However, they suffer from limited diversity and high development costs. In contrast, data-driven methods using deep generative models have emerged as a promising solution, enabling the automatic generation of diverse target scenarios for function development and testing. These methods offer improved scalability and the potential to uncover rare or complex traffic interactions by leveraging large-scale traffic datasets. 

One representative method is the use of Generative Adversarial Networks (GANs), which consist of two main components: a generator and a discriminator. Both are built on deep neural networks and trained in an adversarial process \cite{goodfellow2014generative}. The generator learns to produce realistic data from random noise, while the discriminator learns to distinguish between real and generated samples by evaluating both during training. TrajGAN was proposed to generate lane change scenarios on highways based on the adaption of the InfoGAN \cite{krajewski2018ganvae}. Zhang, et al. \cite{zhang2022drivingbehavior} introduced a scenario generator D2Sim, designed to learn human driving behaviors in mixed traffic environments using Graph Attention Networks with Transformers and GAN. 

Another common deep generative model is the Variational Autoencoder (VAE). A VAE learns a continuous latent representation by optimizing the Evidence Lower BOund (ELBO) of the data likelihood \cite{kingma2014autoencoding}. It consists of an encoder that maps input data to a latent distribution and a decoder that reconstructs samples drawn from this distribution. The incorporation of a KL-divergence term regularizes the latent space, facilitating smooth interpolation and robust generative capabilities across diverse domains. Krajewski, et al. \cite{krajewski2019beziervae} introduced a Bezier layer in decoder of VAE to ensure the smoothness in position and speed domain of trajectories. Jiao, et al. \cite{jiao2022tae} proposed a behavior-aware Trajectory Autoencoder (TAE) that leverages a semi-supervised adversarial autoencoder to encode driver behaviors into the latent space, such as aggressiveness and intention. 

Diffusion Models are generative frameworks that iteratively transform Gaussian noise into structured data by learning the reverse of a predefined forward diffusion process \cite{song2019generative}. This is typically achieved via a parameterized denoising model trained to approximate the posterior distribution of clean data given noisy observations. Guo, et al \cite{guo2023diffusion} proposed a controllable diffusion model for generating driving scenarios, integrating object category and position via bounding boxes. A diffusion-based framework for controllable traffic scenario generation that jointly synthesizes agent poses, orientations, and trajectories was introduced by Pronovost, et al \cite{pronovost2023diffusion}.

However, these methods are insufficient for generating long-term scenarios, often lacking fine-grained trajectory details or focusing on a single agent. Such limitations reduce their applicability in effectively supporting scenario generation for function development and comprehensive testing. Additionally, roundabout scenarios, which present complex multi-agent interactions, remain underrepresented in existing work. To address this gap, the following section presents the proposed CVAE-T model designed for learning and generating multi-agent scenarios in roundabouts.

\section{Methodology}

\subsection{Data Processing}

\begin{figure}[htbp]
\centerline{\includegraphics[width=\linewidth]{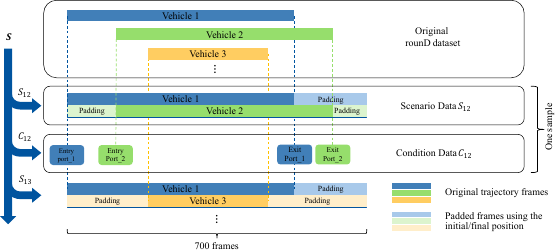}}
\caption{Data processing pipeline for extracting trajectories in the same temporal window}
\label{dataprocess}
\end{figure}

The rounD dataset was released in 2020 as an open-source dataset for non-commercial use. The Neuweiler roundabout in Aachen from the dataset, as shown in Figure \ref{roundabout}, is selected for this study. It is a four-armed roundabout and each entry-exit arm is named from A to D.

In this study, a scenario consists of two vehicle trajectories concurrently navigating the roundabout with spatial or temporal interactions. To improve data quality for training, all vehicle trajectories with fewer than 250 frames are removed. The initial and final frame indices of each vehicle are available in the dataset. Based on this information, two vehicles that appear within the same temporal window are combined together as a scenario. When the temporal overlap between two vehicles is too short, it will be difficult for the model to learn meaningful interactions. Thus, the two trajectories with an overlap length of fewer than 100 frames are excluded from the training data.

Since the trajectory lengths of the two vehicles in each scenario differ, temporal alignment within a common time window is required.  Although generative model can be designed to process variable-length input, this often leads to significantly increased training complexity \cite{petrovich2021action}. As Figure \ref{dataprocess} shows, a fixed sequence length of 700 frames (28 s) is selected to store the trajectories of the two interacting vehicles. From our analysis of the dataset, this duration is suitable for capturing the entire trajectories of both vehicles while maintaining a reasonable data size for efficient training.

The longitudinal positions $x_{vehicle ID}^{frameID}$ and lateral positions $y_{vehicle ID}^{frameID}$ of the two vehicles are extracted and and concatenated to form a matrix \( \mathbf{S} \in \mathbb{R}^{700 \times 4} \) as scenarios:

\begin{equation}
\mathbf{S} = \left( \mathbf{P}_t \right)_{t=1}^{700}, \quad \mathbf{P}_t = \left( x_1^t, y_1^t, x_2^t, y_2^t \right) \in \mathbb{R}^4
\end{equation}

 Each vehicle has four possible entry and four possible exit ports in the roundabout. Although some vehicles enter and exit at the same port or complete more than one full circle, these cases are rare and not representative of typical roundabout behavior. Therefore, these scenarios are excluded from this study. Each vehicle has 12 entry–exit possibilities. Categorical condition data $\mathbf{C}$ is derived from the entry–exit combinations of the two vehicles and the number of unique combinations is computed as:

\begin{equation}
\text{Number of condition categories} = \frac{n(n+1)}{2} \\
\end{equation}

where \( n = 12 \) and the total number of unique condition categories is 78, resulting in \( \mathbf{C} \in \{1, 2, \dots, 78\} \). The scenario data are structured with shape (N, 700, 4) and the condition data have shape (N, 1), where N denotes the number of samples. Categories with fewer than 300 samples are excluded from the training process. As a result, the valid sample size N is 30329, with a total of 28 valid condition categories. To improve training efficiency, the trajectory sequences are downsampled by a factor of 3, reducing the frame length from 700 to 234 frames, resulting in a temporal resolution of $3/25s=0.12s$. 

\subsection{Conditional Variational Autoencoder}\label{CVAE}

Conditional Variational Autoencoder (CVAE) is a generative model that extends the VAE by incorporating additional conditional information \cite{kingma2014semi}. This enables the generation of data $\mathbf{S}_{gen}$ conditioned on known attributes, which in this study correspond to synthetic scenarios consisting of two trajectories in roundabouts. The scenario data $\mathbf{S}$ represent the training data and condition data $\mathbf{C}$ denote the categories of the entry and exit combination. The CVAE introduces a latent variable \( \mathbf{z} \)  to model the underlying structure and the joint conditional generative process is defined as

\begin{equation}
p_\theta(\mathbf{S}, \mathbf{z} \mid \mathbf{C}) = p_\theta(\mathbf{S} \mid \mathbf{z}, \mathbf{C}) \cdot p(\mathbf{z}),
\end{equation}

where \( p(\mathbf{z}) = \mathcal{N}(\mathbf{0}, \mathbf{I}) \) is a standard normal prior. As the true posterior \( p_\theta(\mathbf{z} \mid \mathbf{S}, \mathbf{C}) \) is intractable, CVAE introduces an encoder \( q_\phi(\mathbf{z} \mid \mathbf{S}, \mathbf{C}) \), which approximates it using a variational Gaussian distribution:
\begin{equation}
q_\phi(\mathbf{z} \mid \mathbf{S}, \mathbf{C}) = \mathcal{N}(\mathbf{z}; \boldsymbol{\mu}_\phi(\mathbf{S}, \mathbf{C}), \text{diag}(\boldsymbol{\sigma}^2_\phi(\mathbf{S}, \mathbf{C}))).
\end{equation}

The latent variable \( \mathbf{z} \) is sampled using the reparameterization trick:
\begin{equation}
\mathbf{z} = \boldsymbol{\mu} + \boldsymbol{\sigma} \odot \boldsymbol{\epsilon}, \quad \boldsymbol{\epsilon} \sim \mathcal{N}(\mathbf{0}, \mathbf{I}).
\end{equation}

where $\odot$ denotes element-wise multiplication. The CVAE is trained by maximizing the conditional Evidence Lower Bound (ELBO), which balances reconstruction quality and latent space regularization. This objective is implemented as a loss function composed of a reconstruction loss and a Kullback–Leibler (KL) divergence loss. The reconstruction loss $\mathcal{L}_{\text{rec}}$ measures how closely the decoder’s output $\hat{\mathbf{S}}$ matches the input data $\mathbf{S}$ and is computed using Mean Squared Error (MSE) between reconstructed and the original scenarios.

\begin{equation}
    \mathcal{L}_{\text{rec}} = \| \mathbf{S} - \hat{\mathbf{S}} \|^2,
\end{equation}
where \( \hat{\mathbf{S}} = p_\theta(\mathbf{S} \mid \mathbf{z}, \mathbf{C}) \) denotes the reconstructed scenario from the decoder.

The Kullback–Leibler (KL) divergence term encourages the approximate posterior to remain close to the prior distribution:
\begin{equation}
    \mathcal{L}_{\text{KL}} = \frac{1}{2} \sum_{j=1}^{L} 
    \left( \mu_j^2 + \sigma_j^2 - \log \sigma_j^2 - 1 \right),
\end{equation}
where \( L \) is the dimensionality of the latent space, and \( \mu_j \), \( \sigma_j \) are the output parameters of the encoder.

The total loss function used during training is:
\begin{equation}
    \mathcal{L}_{\text{total}} = \mathcal{L}_{\text{rec}} + \beta \cdot \mathcal{L}_{\text{KL}},
\end{equation}
where \( \beta \) is a tunable weight that controls the influence of the KL divergence. In this study, \( \beta \) is gradually increased from 0.4 to 0.8 during training. This annealing strategy allows the model to focus initially on reconstruction, delaying the regularization pressure on the latent space to promote more stable convergence and a more structured latent representation.

\subsection{Model Structure}

\begin{figure}[htbp]
\centerline{\includegraphics{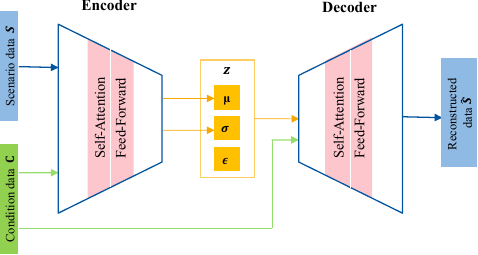}}
\caption{Transformer-enhanced Conditional Variational Autoencoder (CVAE-T) structure}
\label{cvaemodel}
\end{figure}

Transformer was originally proposed for sequence modeling in natural language processing and are capable of handling time-series data \cite{vaswani2017attention}. Furthermore, the integration of Transformer architectures with VAEs has demonstrated strong potential in tasks such as story completion and image processing \cite{wang2019t} \cite{shamsolmoali2024distance}. This approach is also well-suited for multi-agent scenario generation, where vehicle trajectories and interactions rely on spatio-temporal context. Self-attention-based Transformer layers are integrated into the encoder and decoder of the CVAE in this study, to enhance feature extraction from the original scenarios as shown in Figure \ref{cvaemodel}.

The encoder of the CVAE takes the scenario data and condition data as input. Since the condition values range from 1 to 78, directly using these integers as numerical inputs would introduce a false ordinal relationship and mislead model training. To address this, the condition data are first passed through an embedding layer, enabling the model to learn dense semantic representations. These embeddings are then concatenated with the scenario data to form a unified input. A convolutional layer is applied to extract local motion patterns and subsequently, a bidirectional Gated Recurrent Unit (GRU) captures temporal dependencies in both forward and backward directions. The output is then passed through Transformer layers, followed by global average pooling. Finally, fully connected layers generate the parameters \( \boldsymbol{\mu} \)  and \( \log \boldsymbol{\sigma}^2 \) that define the latent distribution, from which the latent variable is sampled using the reparameterization trick.

In the decoder, scenarios are reconstructed from the latent variable $z$ and the conditional input. The latent sequence is first expanded and concatenated with the embedded condition, then passed through a GRU layer to capture initial temporal dependencies. The same Transformer layers used in the encoder are applied in the decoder to further capture global spatio-temporal patterns across the sequence. Finally, a dense layer outputs continuous trajectory features at each time step.

\subsection{Model Configuration and Training}

A total of 70\% of the scenarios are randomly selected as training data, 15\% are used for validation during training, and the remaining 15\% are reserved for testing.  The $\beta$ value is linearly increased from 0.4 to 0.8 over a warm-up period of 200 epochs. The model is trained using the Adam optimizer with a learning rate of $1 \cdot 10^{-4}$. The dimensionality of the latent space is set to 20 to ensure sufficient capacity for capturing relevant scenario features while maintaining effective influence of latent variables on the output. For the Transformer architecture, we use a head size of 256, a feed-forward dimension of 512, and 4 attention heads in multi-head attention layer. The model is trained for 400 epochs with a batch size of 32.

\section{Result}\label{result}

\subsection{Reconstruction and Generation}

In this section, we analyze the reconstruction performance by comparing the original and reconstructed scenarios. Root Mean Squared Error (RMSE) is implemented here to evaluate the reconstruction performance. As shown in Table \ref{vae_rmse}, the RMSE value indicate that the reconstructed trajectories closely follow the original trajectories over the 28 s duration in a scenario. Given the complexity of roundabout scenarios and the extended temporal span in certain conditions, the RMSE values are considered reasonable. The slight reconstruction error is expected due to the reparameterization trick and the increasing $\beta$ value, both of which introduce controlled stochasticity to promote a smooth and generalizable latent space, though at the cost of slightly higher RMSE value.

\begin{table}[h!]
\centering
\caption{Reconstruction performance of proposed CVAE-T model}
\begin{tabular}{|l|c|c|c|}
\hline
\textbf{RMSE} & \textbf{Vehicle 1} & \textbf{Vehicle 2} & \textbf{Total} \\
\hline
Longitudinal positions (m) & 2.3898 & 2.2864 & 2.3387 \\
\hline
Lateral positions (m)      & 1.6851 & 1.9027 & 1.7972 \\
\hline
\end{tabular}
\label{vae_rmse}
\end{table}

\begin{figure}[htbp]
    \centering
    \includegraphics[width=\linewidth]{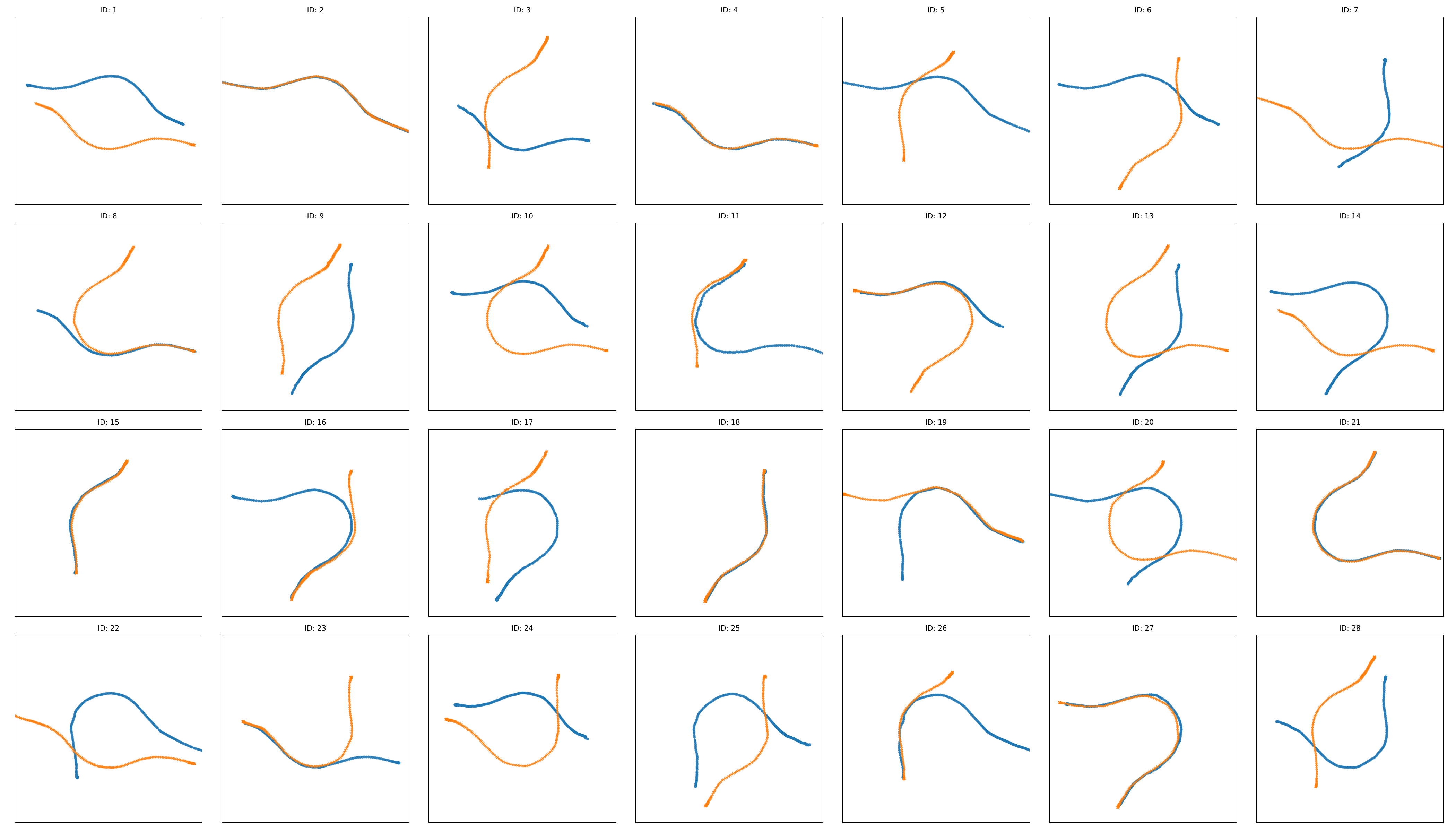}
    \caption{Visualization of generated roundabout scenarios under different conditions}
    \label{gen_samples}
\end{figure}

\begin{figure}[htbp]
    \centering
    \includegraphics[width=\linewidth]{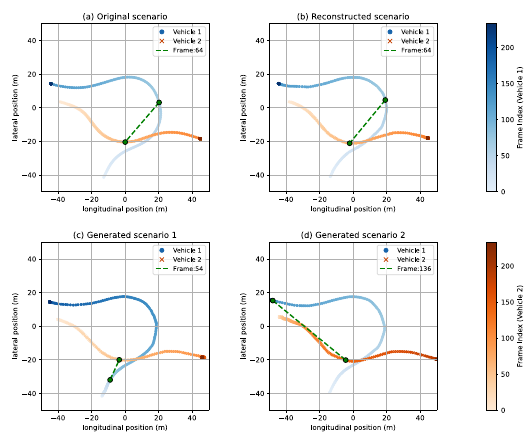}
    \caption{Comparison of scenarios under the same condition}
    \label{compare}
\end{figure}

Twenty-eight scenarios generated under different conditions using the trained model are shown in Fig.~\ref{gen_samples}. The curvature of the generated trajectories indicates that the vehicles are able to successfully navigate the roundabout. For further analysis, one representative condition category is investigated in detail, as shown in Figure~\ref{compare}. In this case, Vehicle 1 (blue) travels from Port C to Port B, while Vehicle 2 moves from Port B to Port D. As the scenario includes both spatio-temporal data of the two vehicles, color transitions from light to dark are used to represent the progression of time.

By comparing the original scenario and reconstructed scenario, the spatio-temporal patterns are shown to be highly consistent. Additionally, the same condition is used to generate synthetic scenarios, as illustrated in Figure~\ref{compare}(c) and (d). In the generated scenario, both vehicles enter and exit the roundabout as expected. The two generated scenarios demonstrate that the model can generate structurally similar scenarios while preserving diversity. Vehicles 1 and 2 start and end at slightly different positions in the two generated scenarios. The trajectory of Vehicle 2 in Figure \ref{compare}(c) is generally lighter in color than in Figure \ref{compare}(d), indicating that it enters and exits the roundabout earlier. In contrast, Vehicle 1 shows the reverse behavior, entering and exiting the roundabout later. The green dashed line indicates the frame when Vehicle 2 enters the conflict area (within a threshold of 5 meters), along with the corresponding position of Vehicle 1.

\subsection{Key-Performance-Indicators}

Key-Performance-Indicators (KPIs) are commonly used to describe the dynamic relationship between two vehicles and to classify the risk level of traffic situation based on predefined thresholds \cite{li2024roundabouts}. Time to Collision (TTC) estimates the remaining time before a collision would occur if both vehicles maintain their current speed and direction. Post Encroachment Time (PET) is another suitable KPI for scenarios in roundabouts, defined as the time gap between one vehicle leaving and another entering a shared conflict zone where their paths intersect. For both TTC and PET, lower values indicate more critical scenarios.

\begin{figure}[htbp]
    \centering

    \begin{subfigure}{\linewidth}
        \centering
        \includegraphics[width=\linewidth,height=0.45\textheight,keepaspectratio]{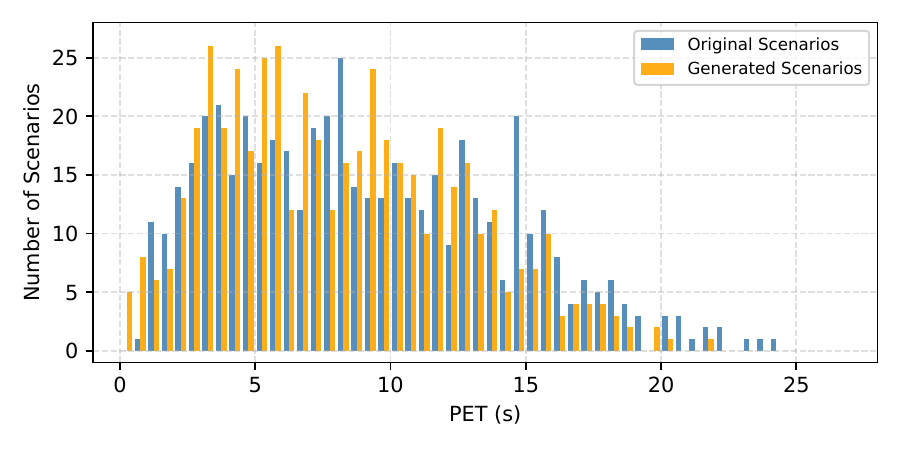}
        \caption{Distribution of PET in original and generated scenarios}
    \end{subfigure}

    \vspace{1em}

    \begin{subfigure}{\linewidth}
        \centering
        \includegraphics[width=\linewidth,height=0.45\textheight,keepaspectratio]{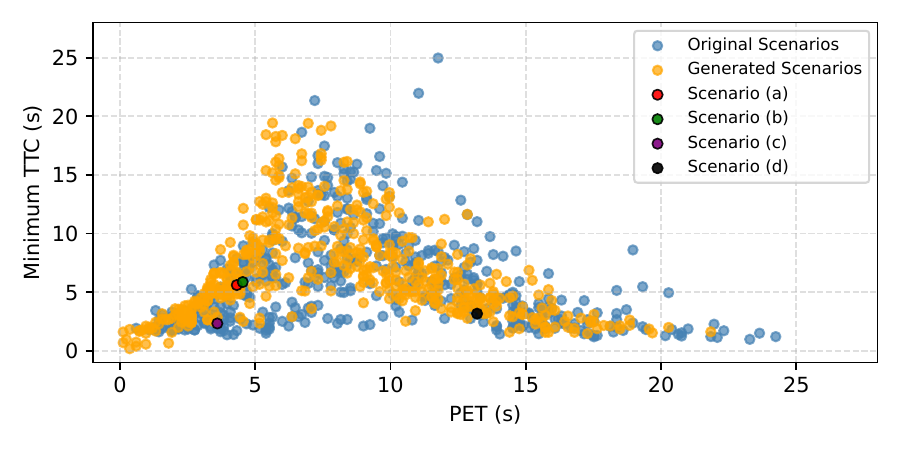}
        \caption{Scatter plot of PET against minimum TTC in original and generated scenarios}
    \end{subfigure}

    \caption{Comparison of 500 original and 500 generated scenarios based on KPIs}
    \label{petttc}
\end{figure}

Randomly selected 500 original scenarios from the dataset and 500 generated scenarios under the same condition are compared based on TTC and PET values in Figure \ref{compare}. TTC is computed at each frame, and the minimum value within the scenario is used as the representative value. In the scenarios, a conflict zone is always present, ensuring the availability of a valid PET value. Trajectories outside the roundabout are not the focus of this work and are therefore excluded from the analysis. As shown in Figure \ref{petttc}(a), PET values in the original scenarios predominantly range from 1.5 s to 16 s. The generated scenarios exhibit a similar distribution, although with a higher number of critical cases where PET is less than 1.5 s. Figure \ref{petttc}(b) presents a scatter plot of PET against the corresponding minimum TTC values for each scenario. The observed distribution indicates that the model is capable of capturing vehicle interactions by replicating realistic patterns in both KPIs.

In addition, the TTC and PET values of the 4 scenarios in Figure~\ref{compare} were computed and visualized in the scatter plot. Compared to the original scenario \ref{compare}(a), Vehicle 1 enters the roundabout later in generated scenario \ref{compare}(c), resulting in both vehicles having a higher likelihood of collision within the conflict zone. As a result, small PET and minimum TTC values are observed near the conflict zone. In contrast, in generated scenario \ref{compare}(d) Vehicle 2 enters the roundabout significantly later, while Vehicle 1 is already approaching Port B to exit. This lead to a large PET value, while minimum TTC is reached near Port B with small value. 

\subsection{Latent Space Disentanglement and Interpretability}\label{AAA}

In roundabouts, the entry and exit timing as well as velocity profiles of the interacting vehicles are key factors influencing the safety level of the scenarios. Therefore, we investigate whether the proposed CVAE-T model captures such parameters or other relevant parameters. We manually vary one latent dimension from -3 to 3 in steps of 1.5, while keeping all other latent variables fixed at zero. Scenarios are subsequently generated based on these controlled latent inputs. In addition, velocity profiles are computed from position data, excluding segments outside the roundabout.

\begin{figure*}[htbp]
    \centering

    \begin{subfigure}{\linewidth}
        \centering
        \includegraphics[width=\linewidth,height=0.45\textheight,keepaspectratio]{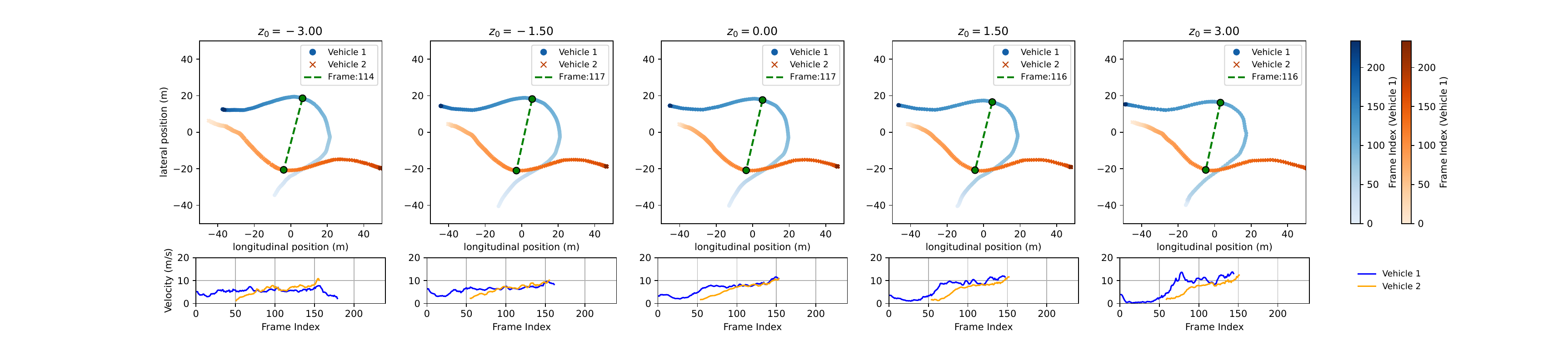}
        \caption{Variation of 1\textsuperscript{st} latent parameter}
    \end{subfigure}

    \begin{subfigure}{\linewidth}
        \centering
        \includegraphics[width=\linewidth,height=0.45\textheight,keepaspectratio]{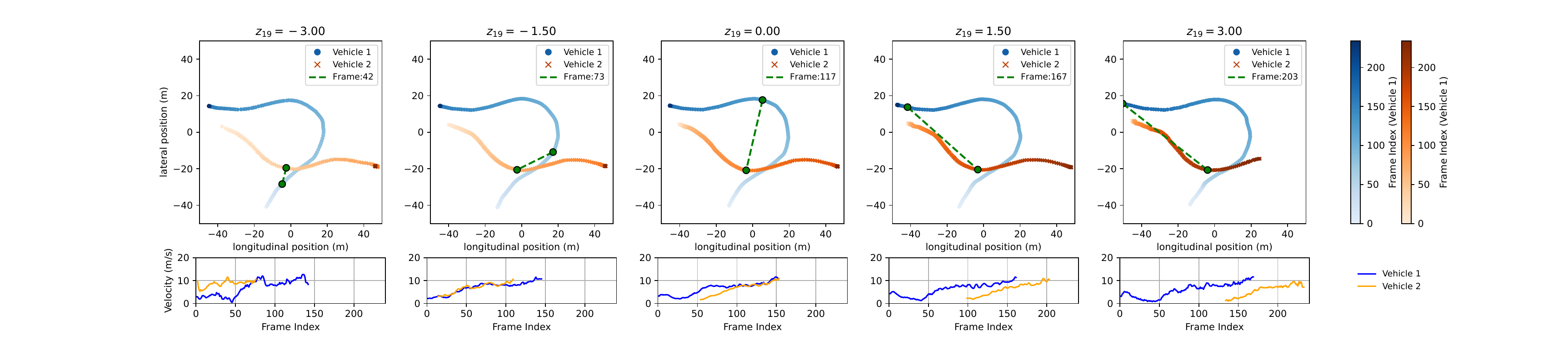}
        \caption{Variation of 20\textsuperscript{th} latent parameter}
    \end{subfigure}

    \caption{Influence of latent parameter variation on generated scenarios}
    \label{latentpara}
\end{figure*}

Five latent parameters exhibit a distinct influence on the interactive behavior between the two vehicles in the generated scenarios. Two representative latent parameters are here analyzed to explore their influence on scenario generation. Figure \ref{latentpara}(a) illustrates the effect of varying the 1\textsuperscript{th} latent parameter from -3 to 3 (left to right). While the trajectory shape of Vehicle 1 remains relatively constant, its velocity profile changes noticeably, particularly at the roundabout entry. In the first scenario, the velocity is steady, whereas in the fifth, the vehicle decelerates before entering the roundabout and accelerates after entry. Figure \ref{latentpara}(b) shows the effect of manipulating the 20\textsuperscript{th} latent parameter. As its value increases, the entry timing of Vehicle 2 is delayed, though the overall velocity profile remains similar.

This demonstrates the model's strong potential for generating diverse scenarios, such as those involving different entry and exit timings or varying velocity profiles.

\section{Conclusion}

In this paper, we investigate multi-agent scenario generation in roundabouts. The rounD dataset is selected and scenarios involving two interacting vehicles within the same temporal window are extracted. A CVAE is employed to learn the trajectories of the two vehicles and their interactions. To enhance the model’s ability to capture long-term spatio-temporal patterns, Transformer layers are integrated into the CVAE model.

The results demonstrate that the proposed model is capable of reconstructing original scenarios and generating synthetic scenarios based on given conditions. Two KPIs, namely TTC and PET are employed to evaluate generation performance by comparing their distributions against those from the original dataset. Furthermore, the disentanglement and intuitiveness of the latent space are analyzed. Two latent dimensions are selected, each demonstrating a clear relationship with the vehicle's velocity and roundabout entry timing, respectively. These findings highlight the ability of the proposed model to augment existing datasets with diverse and safety-relevant scenarios. In particular, it enables the generation of condition-specific scenarios tailored for the training and evaluation of intelligent driving functions such as AEB, FCW and CA.

While the proposed model is capable of generating most roundabout scenarios realistically, performance degrades in conditions with insufficient training samples. Moreover, the proposed CVAE-T model relies solely on position data, which limits its ability to generate scenarios with smooth velocity profiles. Future work will incorporate vehicle acceleration profiles, vehicle geometry and environmental elements, such as road boundaries, to enable the generation of more realistic edge-case scenarios.

\bibliographystyle{IEEEtran}
\bibliography{refs}

\end{document}